# Some Ethical Issues in the Review Process of Machine Learning Conferences

**Alessio Russo**

Division of Decision and Control Systems, EECS School, KTH Royal Institute of Technology, Sweden

*Abstract-* Recent successes in the Machine Learning community have led to a steep increase in the number of papers submitted to conferences. This increase made more prominent some of the issues that affect the current review process used by these conferences. The review process has several issues that may undermine the nature of scientific research, which is of being fully objective, apolitical, unbiased and free of misconduct (such as plagiarism, cheating, improper influence, and other improprieties). In this work, we study the problem of reviewers' recruitment, infringements of the double-blind process, fraudulent behaviors, biases in numerical ratings, and the appendix phenomenon (i.e., the fact that it is becoming more common to publish results in the appendix section of a paper). For each of these problems, we provide a short description and possible solutions. The goal of this work is to raise awareness in the Machine Learning community regarding these issues.

## I. INTRODUCTION

With this essay, we wish to reflect on some ethical issues that regard the review process of some of the main Machine Learning (ML) conferences. ML has seen a sharp increase in interest during the last decade, due to the successes of Deep Learning [1]. This interest has led to a steep increase in the number of papers being submitted to conferences such as NeurIPS, ICML, ICLR, and so on. As a comparison, between 2014 and 2020 the number of submissions at NeurIPS have increased at an average of +34% submitted papers every year [2,3]. However, despite a larger body of work, the acceptance rate of these conferences has roughly remained the same over the years [2,3]. This point seems to suggest that these conferences scale well with the number of submissions. Nonetheless, based on some facts that we discuss in this work, the review process has several issues that may undermine the nature of scientific research, which is of being fully objective, apolitical, unbiased and free of misconduct (such as plagiarism, cheating, improper influence, and other improprieties).

Before delving into the topic of this paper, we first briefly discuss in Section (2) what is the *de facto* review process in the largest ML conferences. Then, in Section (3), we highlight some of the problems in the review process and discuss possible solutions. Finally, we conclude in Section (4).

## II. BACKGROUND: REVIEW PROCESS

The review process in large-scale ML conferences can be summarized as follows: it is a double-blind process that includes one rebuttal period. The review process lasts roughly 9-10 weeks, followed by a rebuttal period that lasts 1 week in which the authors can reply to the reviewers. Reviewers choose the papers to review according to a bidding system or review a certain paper upon invitation of an Area Chair (AC). The number of reviewers per paper varies, but on average it is 3-4, which are supervised by an AC. Each reviewer provides a numerical score for the paper, and also a confidence score. Depending on the conference, the authors may have the possibility to reply to single reviewers (e.g., in https://openreview.net/), or can just reply to all the reviewers in 1 PDF page (this is the case for NeurIPS). Roughly 2-3 weeks after the rebuttal period, the authors receive a final decision (acceptance/rejection) that is made by the AC according to the a *posteriori* rebuttal reviews.

## III. DISCUSSION

Based on the figure that the acceptance rate has roughly remained the same over the years, one may argue that ML conferences have scaled well to the increasing amount of work being done by the ML community. However, the previous statement is false in case these conferences did not manage to maintain the same *quality standards* over the years. By quality standards, we mean a review process that can create scientific knowledge through qualified competence that adheres to the values of scientific research (values that we mentioned in the introduction).

However, as we will soon point out in the next paragraphs, some of the problems are not caused by having a larger number of submitted papers. Some of them just became more evident because of this larger body of work.
As a matter of fact, recently many researchers have raised concerns in regards to the current review process, and some of them have advocated for profound changes in the review system [4-10].

We now present some of the main ethical issues in the review process, without any order of priority. These issues are:
- The recruitment of reviewers, and lack of reviewing experience
- Infringements of the double-blind process
- Fraudulent behavior in bidding systems
- Biases in numerical ratings
- The appendix phenomenon

*A. Recruitment of reviewers, and lack of reviewing experience*

Being able to recruit reviewers plays a large role in scaling up conferences. Conferences, and also journals, rely on the community to thrive. However, conferences do not hire, and do not pay, reviewers (even though the same reviewers may need to pay to publish a paper/go to the conference). As a consequence, the number of papers to review for each reviewer should be limited. If the number of submitted papers increases and the length of the review process remains fixed, it follows naturally that an increase in the number of submitted papers must be followed by a proportional increase of reviewers. Moreover, we also need to consider the event that some reviewers may *retire* (i.e., just do not want to review anymore): this exacerbates the effect of requiring additional reviewers each year.

Therefore, as we have just seen, an increase in the number of submitted papers requires a proportional increase in the number of new reviewers. However, as pointed out in [6-8], the number of qualified reviewers is growing at a much slower rate. In [8] the president of the ICML board John Langford says that "There is significant evidence that the process of reviewing papers in machine learning is creaking under several years of exponentiating growth".

Conferences to cope with this sharp increase have lightened some of the requirements needed to become a reviewer. In NeurIPS 2016 36% of the reviewers were recruited by asking the authors to nominate at least one more author who is willing to become a reviewer [6,9]. However, a large fraction (7 out of 10) of the new reviewers consisted of Ph.D. students, i.e., junior researchers that may not have the knowledge or the experience to objectively evaluate new scientific research [6,9]. Sometimes, new reviewers may even be master students. Recently, ICML 2020 made a public call for reviewers from self-nominated individuals [6].
Overall, we can conclude that this type of action increases the likelihood of poor-quality reviews, with the effect of producing a sense of *randomness* in the review process [10].

Unfortunately, it is not easy to come up with a remedy for the lack of qualified reviewers. A long-term action would be to include in the scholars' curricula a mandatory course on the review process. This is partly done in many universities where the students are asked to evaluate each other, but, unfortunately, this is not always the case for Ph.D. students. A similar suggestion has been recently proposed in [4], where it has been suggested to include mentorship in reviewing. As a matter of fact, a recent workshop conducted at NeurIPS 2020 seems to indicates the effectiveness of mentoring scholars into producing high-quality reviews.

*B. Infringements of the double-blind process*

Most conferences, if not all, allow authors to publish pre-prints on public websites such as arXiv, etc…. Sometimes authors may even publish the code of their algorithm on public repositories (e.g., in GitHub). As a consequence, these actions break the double-blind review process.

Breaking the double-blindness property induces bias in the review process. In [12] they found statistical evidence of a positive correlation between acceptance rate and papers with high reputation released on arXiv, where reputation is based on the number of Google Scholar citations. Moreover, always in [12], they point out that less confident reviewers may be prone to give high scores to well-known authors, and low scores to less-known authors. Although their analysis does not show causation, it shows that there is indeed a positive correlation in place.

The issue could be partially solved if websites such as arXiv permit authors to anonymously publish their work, as proposed in [12]. In extreme cases, one can also forbid publications of pre-prints. This ban may reduce the velocity at which new knowledge is disseminated. However, this would result in a pure double-blind review process that could help to reduce the bias.

*C. Fraudulent behavior in bidding systems*

Large-scale ML conferences nowadays employ a bidding system to let reviewers choose the paper they want to review. However, a bidding system may lead to agreements between reviewers (*quid-pro-quo* arrangements) with the aim of providing positive, or negative, reviews. Similarly, reviewers may also try to deliberately reject papers that they dislike for their own benefit.

This behavior, unfortunately, is not rare in academia. On June 13 2019, Huixiang Chen, a Ph.D. student whose paper was about to be published at ISCA 2019 in Pheonix, hanged himself in the campus building of the University of Florida [13,14]. The student left a note, saying he refused to continue committing acts of academic dishonesty and accused his advisor Dr. Tao Li [14]. A Joint ACM-IEEE Investigative Committee has launched an investigation, determining that "*that there was no evidence of misconduct as part of the paper review process.*" [14]. However, as pointed out in [13], the fact that Huixiang's laptop contained most of the submissions to more than one conference, along with the de-anonymized reviews and discussions, is disturbing and troublesome. In [13] they always point out that "*another ACM SIG community had a collusion problem where the investigators found that a group of PC members and authors colluded to bid and push for each other's papers violating the usual conflict-of-interest rules*" [13].

In [15] Langford raises the issue of *torpedo reviewing*, that is "*if a research direction is controversial in the sense that just 2-or-3 out of hundreds of reviewers object to it, those 2 or 3 people can bid for the paper, give it terrible reviews, and prevent publication. Repeated indefinitely, this gives the power to kill off new lines of research to the 2 or 3 most close-minded members of a community, potentially substantially retarding progress for the community as a whole.*"

Always in [15] Langford points out that torpedo reviewing may not be a fantasy. Sometimes people tend to bid on papers they want to reject "*on the theory that rejections are less work than possible acceptances*" [15].

Another issue is that of *rational cheating* [17], where reviewers tend to reject papers that compete with their own work. In [16] they analyze the strategic behavior of people in competitive peer-reviews, suggesting that "*competition incentivizes reviewers to behave strategically, which reduces the fairness of evaluations and the consensus among referees.*"

All these behaviors seem to be part of human nature. As humans, we tend to prioritize our survival chances. Therefore, one may argue that humans, by nature, tend to prefer individual gains to collective gains, leading to the creation of malicious patterns in organized small groups of individuals.

To solve these issues conferences may employ randomization in the review process. Randomizing the reviewers would help to reduce the likelihood of a malicious reviewer getting the paper he/she wants to review. However, randomization may lead to lower quality in the review process, since authors would not be able to bid papers for which they are competent. This solution is further investigated in [11], where the authors propose an algorithm to randomize the review process.

### *D. Biases in numerical ratings*

The reviewers in their reviews have to provide a numerical rating for each paper they review, together with a confidence score. It is now widely known that mapping opinion to a single number is a difficult operation for people [18].

Moreover, as pointed out in [20], another problem of numerical ratings is that of *selection bias* [19]. Selection bias refers to the phenomenon that leads average ratings to have a positively skewed tendency. The idea is that "*users are likely to select already top-ranked entities to expression opinion on and rate*" [20]. This agrees with the observation that there is a positive correlation between acceptance rate and preprints with a high reputation [12], as previously discussed in section 3.B.

In the review process, reviewers may be influenced by the attitude/comments of other reviews during the rebuttal period. This is also called *conformity bias*, which is the desire to conform to a group opinion [21-23], which can persist even in presence of overwhelming contrary evidence [23]. Consequently, less confident reviewers may be induced into conforming to the opinion of more confident reviewers.

Instead of providing a numerical rating, reviewers may be asked to make a comparison or a ranking, as suggested in [18]. To address the problem of conformity bias, one may use the idea of having multiple groups of reviewers, where each group is blind to each other, review the same paper. Reviews may be then randomized, and the AC will take a decision based on these randomized reviews.

### *E. The appendix phenomenon*

Last, but not least, a critical issue in ML conferences is that a major part of the results appears in the appendix, or supplementary material [4]. This is what we call the *appendix phenomenon*. Conferences have usually been the venue for fast dissemination of research, but the sharp increase in the number of papers has led towards an arms race where the authors tend to make more comprehensive papers at the cost of abusing the appendix. This may lead the reviewers to ask for more work since the bar for comprehensiveness keeps getting raised [4].

Moreover, it is common knowledge that most of the reviewers tend not to review the supplementary material. The fact that most of the results not get reviewed poses a great threat to the dissemination of knowledge and the development of the ML community.

A quick and rapid solution would be to limit the number of pages submitted to a conference, or limit the number of pages of an appendix. Authors should be able to state their results in a limited number of pages. If this is not possible, then the authors should consider submitting the paper to a journal.

## IV. CONCLUSION

In this work, we have analyzed some of the ethical problems that currently exist in the review process of some of the most known Machine Learning conferences. These problems are mostly related to the lack of qualified reviewers, the infringements of the double-blind process, the fraudulent behavior of the bidding system in the review process, the human bias in giving a numerical rating, and, the fact that most of the results appear in the supplementary material of a paper.

Some of these issues can be solved by using the following measures: (1) introduce an appropriate training/mentorship for new scholars, (2) randomizing the review process, (3) anonymizing preprints, (4) use rankings instead of numerical ratings, (5) having multiple groups of reviewers that are blind to each other, (6) limit the number of pages of an appendix, or completely remove the possibility of uploading supplementary material.

All these issues deserve more space than what was given here. However, the goal of this work is to highlight some of the issues the ML community is currently facing, and raise the awareness of researchers to these problems.

AUTHORS

**Alessio Russo** – Ph.D. Student at the Division of Decision and Control Systems, KTH Royal Institute of Technology (alessior@kth.se)